\documentclass[10pt,twocolumn,letterpaper]{article}

\usepackage{wacv}
\usepackage{times}
\usepackage{epsfig}
\usepackage{graphicx}
\usepackage{amsmath}
\usepackage{amssymb}
\usepackage{booktabs}
\usepackage{multirow}

\usepackage[linesnumbered,ruled]{algorithm2e}
\usepackage{algpseudocode}
\usepackage[utf8]{inputenc} 
\usepackage[T1]{fontenc}    
\usepackage{url}            
\usepackage{amsfonts}       
\usepackage{nicefrac}       
\usepackage{microtype}      
\usepackage{xcolor}         
\usepackage{wrapfig}
\usepackage{multicol}

\usepackage{bm}
\usepackage{bbm}
\usepackage{mathtools}
\usepackage{listings}
\usepackage{xcolor,colortbl}

\usepackage{comment}
\definecolor{Gray}{gray}{0.85}
\definecolor{LightCyan}{rgb}{0.85,0.92,0.90}

\wacvfinalcopy 

\ifwacvfinal
\usepackage[pagebackref=true,breaklinks=true,colorlinks,bookmarks=false]{hyperref}
\else
\usepackage[pagebackref=true,breaklinks=true,colorlinks,bookmarks=false]{hyperref}
\fi

\begin{document}

\title{Joint Debiased Representation and Image Clustering Learning \\ with Self-Supervision\vspace{-10pt}}

\author{
Shunjie-Fabian Zheng\textsuperscript{\rm 1}\thanks{Equal Contributions.}, 
JaeEun Nam\textsuperscript{\rm 1}\footnotemark[1],
Emilio Dorigatti\textsuperscript{\rm 1},
Bernd Bischl\textsuperscript{\rm 1}, 
Shekoofeh Azizi\textsuperscript{\rm 2}\footnotemark[2],
Mina Rezaei\textsuperscript{\rm 1}~\thanks{mina.rezaei@stat.uni-muenchen.de, shekazizi@google.com}\\
 \textsuperscript{\rm 1} Department of Statistics, LMU Munich, Germany\\
 \textsuperscript{\rm 2} Google Research, Brain Team, Canada\\
}

\maketitle
\thispagestyle{empty}

\begin{abstract} \label{sec:abstract}
\vspace{-8pt}
Contrastive learning is among the most successful methods for visual representation learning, and its performance can be further improved by jointly performing clustering on the learned representations.
However, existing methods for joint clustering and contrastive learning do not perform well on long-tailed data distributions, as majority classes overwhelm and distort the loss of minority classes, thus preventing meaningful representations to be learned.
Motivated by this, we develop a novel joint clustering and contrastive learning framework by adapting the debiased contrastive loss to avoid under-clustering minority classes of imbalanced datasets.
We show that our proposed modified debiased contrastive loss and divergence clustering loss improves the performance across multiple datasets and learning tasks.
The source code is available at \url{https://anonymous.4open.science/r/SSL-debiased-clustering}

\end{abstract}

\vspace{-12pt}
\section{Introduction}
\vspace{-6pt}
\begin{figure*}[!t]
    \centering
    \includegraphics[width=0.99\textwidth]{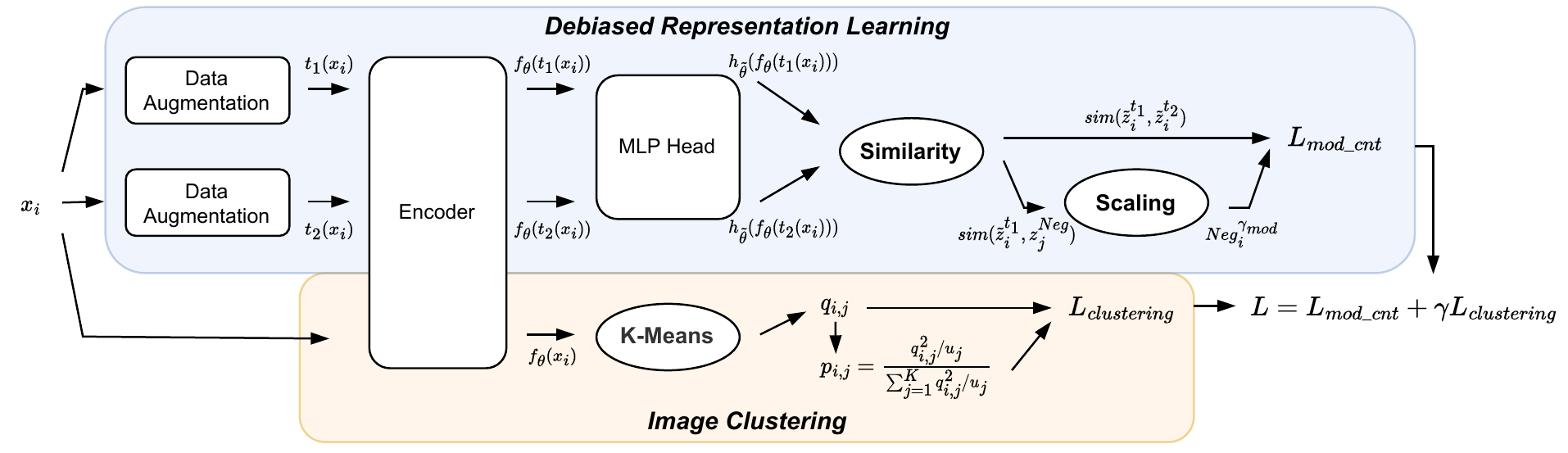}
    \caption{Illustration of the proposed unsupervised debiased representation learning framework. Our method is composed of two parallel deep convolutional transformer architecture where the clustering network takes an original images and the representation network takes two augmented views of the image. The representation network first projects the augmented views onto an embedding space and then processes these representations in a MLP head, which generates the baseline for the pair-wise contrastive objective. Here, we scale the negative sampling strategy of ~\cite{chuang_debiased_2020} by exponential weighting in order to create an excess of debiased negative samples that leads to under-clustering. The clustering network uses the extracted features from the encoder and employ a K-Means clustering with a KL-divergence loss with the students t-distribution as the soft assignments and a target distribution, as a function of the soft assignments~cite{xie2016unsupervised}.}
    \label{fig:method}
\end{figure*}

Self-supervised learning (SSL) has achieved superior performances and outperformed supervised learning models in different research areas such as computer vision~\cite{chen2020big,chen_simple_2020}, natural language processing~\cite{devlin_bert_2019}, and more recently medical image analysis~\cite{azizi2022robust,azizi2021big,taher2022caid} and bio-informatics~\cite{gunduz2021self}.
SSL algorithms learn representations from large scale unlabeled data by solving a \textit{pretext task} such as solving jigsaw puzzles~\cite{noroozi2016unsupervised}, predicting geometric transformations~\cite{hashim2022transformnet}, Bregman divergence learning~\cite{rezaei2021deep}, predicting reverse-complement of genome sequences~\cite{gunduz2021self}, the relative positioning of patches~\cite{doersch2015unsupervised}, etc.
The representations learned by performing such task can then be used as a starting point for different \textit{downstream} tasks such as classification~\cite{van_gansbeke_scan_2020}, semi-supervised learning~\cite{bai2021self}, clustering~\cite{park_improving_2021}, or image generation~\cite{hinz2018image}.
The performance of self-supervised learning was recently further improved by contrastive methods that train a network to maximize the similarity of representation obtained from different augmented views of the same image~\cite{henaff2020data,oord2018representation,he_momentum_2020,chen_simple_2020,chen2020big}.

Recent studies have shown that self-supervised pretext task learning benefits from multi-task learning~\cite{doersch2017multi,rezaei2021learning} such as performing clustering on the learned representations~\cite{zhang_supporting_2021}. 
However, in spite of the explicit clustering, representations learned in such a way could still exhibit overlap between different classes, particularly for complex datasets with a large number of categories and long-tailed data distribution~\cite{liu_large-scale_2019}.
There are three reasons underlying this phenomenon.
First, \cite{chuang_debiased_2020} showed how the representations induced by traditional contrastive learning are inherently biased as each augmentation is contrasted against \emph{all} other samples in the batch, including potentially those of the same underlying latent class.
Second, clustering techniques based on euclidean metrics such as K-means~\cite{macqueen_methods_1967} or Gaussian mixture models ~\cite{celeux_gaussian_1995} struggle when operating on data lying on high-dimensional manifolds~\cite{holte1989concept,rezaei2020statlearning,soleymani2022deep}.
Finally, especially in imbalanced datasets, the loss of samples belonging to minority classes is distorted when their representations are contrasted with an excessive number of negative samples from other classes.
This limits the learning signal to the network and prevents it from assigning similar representations to minority samples while keeping them sufficiently distinct from those of all the other samples.

In this paper, we thus improve simultaneous clustering and contrastive representation learning for imbalanced datasets by generalizing the debiased contrastive loss of \cite{chuang_debiased_2020} to avoid under-clustering minority classes.
An excessive amount of negative samples forces the formation of different clusters within the same category~\cite{wang_solving_2021}, therefore we modified that loss with a smoothing terms that controls the influence of the contrasted negative samples, preventing the previously mentioned phenomenon of under-clustering.
We then use these representations to directly perform clustering~\cite{xie2016unsupervised}.
Our main contributions can be summarized as:


\begin{itemize}
    \item We propose a joint framework for self-supervised learning of visual representations and image clustering. Our proposed method learns debiased contrastive visual representation and unsupervised clustering using divergence loss over the data distributions.
    \item We show empirical results to highlight the benefits of avoiding under-clustering while learning representations using a multi-task learning loss. The model is able to distinguish distinct classes better or at least comparable to the state-of-the-art methods for several benchmark datasets and public medical datasets, characterized by long tailed distributions. Our method is evaluated in linear, semi-supervised, and unsupervised clustering settings on public datasets, achieving comparable or higher performance in comparison to state-of-the-art contrastive learning methods in several tasks. 
\end{itemize}

\section{Related Work} 

\paragraph{Self-supervised Learning}
Self-supervised learning methods have made a great impact in recent years. Initial studies in self-supervised representation learning focused on the problem of learning embeddings without labels such that a linear classifier on the learned embeddings could achieve competitive accuracy as supervised models~\cite{doersch2015unsupervised,rezaei2020generative,rezaei2018generative}. Furthermore, the trained embeddings are also transferable to other tasks.
Early methods explore the inter-class structure of the data during pretraining task by generating the masked tokens~\cite{devlin_bert_2019} and future tokens~\cite{radford_improving_2018} over denoising corruptions~\cite{lewis_bart_2020} to the prediction of the colorization of images~\cite{zhang_colorful_2016}.
In this sense SSL tries to substitute the ground-truth semantic labels with a clever use of data augmentation techniques~\cite{van_gansbeke_scan_2020,rezaei2021deep}, takes replaces of the label signal in a supervised setting in order to solve the pretext task. These augmentations range from colorization~\cite{zhang_colorful_2016}, over rotational predictions~\cite{gidaris_unsupervised_2018} to random cropping~\cite{van_gansbeke_scan_2020}. The aim of the supervision via data augmentation (hence the name Self-supervision) is to mimic prior knowledge through linear constraints derived from the data structure itself. 
Instance-wise constrastive learning treats each data point and its transformations as a separate class and pulls the embeddings of a same class close to each other, while pushing the embeddings of different classes far away~\cite{van_gansbeke_scan_2020}. Hence, contrastive learning can be interpreted as a self-supervised Version of the triplet loss. The primary focus of the research of contrastive losses are grounded in different approaches of generating the positive pairs in vision tasks~\cite{chen_simple_2020,van_gansbeke_scan_2020,rezaei2021deep}. Only in the last two years has the perspective changed and ~\cite{chuang_debiased_2020} approaches the sampling problem of treating random data points as dissimilar, even though they might as well belong to the same class. The correction for same label data points within the regular treatment of such as negative examples culminates to the debiased contrastive loss. ~\cite{robinson_contrastive_2020} introduces a control element to the \emph{hardness} of negative examples, as identifying negative examples close to the decision boundary can improve the training performance. 
We take the negative sampling strategy of the debiased contrastive loss and introduce under-clustering, which happens in cases of an excessive amount of negative samples.

\paragraph{Unsupervised Image Clustering}
Unsupervised clustering denotes a set of techniques that has the goal of finding structure within the latent space in order to group data points with similar traits~\cite{park_improving_2021}. As real world data is frequently high dimensional as in pixels per image and long texts, unsupervised clustering algorithms conduct a dimensionality reduction, which extracts lower dimensional features from the data with no label information. The learned features in the latent space are then subject to a boundary identification over a similarity metric~\cite{sungwonhan}. Among the most popular partition methods are K-Means, which assumes $k$ different classes and assigns the data points to the class, whose centroid is closest to the data point in the latent space~\cite{macqueen_methods_1967}.
{~\cite{van_gansbeke_scan_2020} proposed a multi-step approach called SCAN, in which high level feature embeddings are generated by representation learning and then used as a baseline for the subsequent clustering task via nearest neighbors. Following SCAN ~\cite{park_improving_2021} developed the so-called RUC algorithm~\cite{park_improving_2021}, whereby an add on module is used on top of a generic unsupervised clustering method in order to improve the performance. It is first assumed that the clustering results are noisy as assigned labels can be false. The correctly assigned data points are selected to form a labeled dataset while the other samples are considered unlabeled, which allows the subsequent network to learn in a semi supervised manner with label smoothing\cite{niu2021spice}. Most recently, SPICE~\cite{niu2021spice} was proposed, an algorithm consisting of two semantic aware pseudo labeling networks to allow for reliable self supervision via self-labeling and the extraction of semantically crucial features reflecting the ground truth.}

\begin{figure*}[!t]
    \centering
    \includegraphics[width=0.89\textwidth]{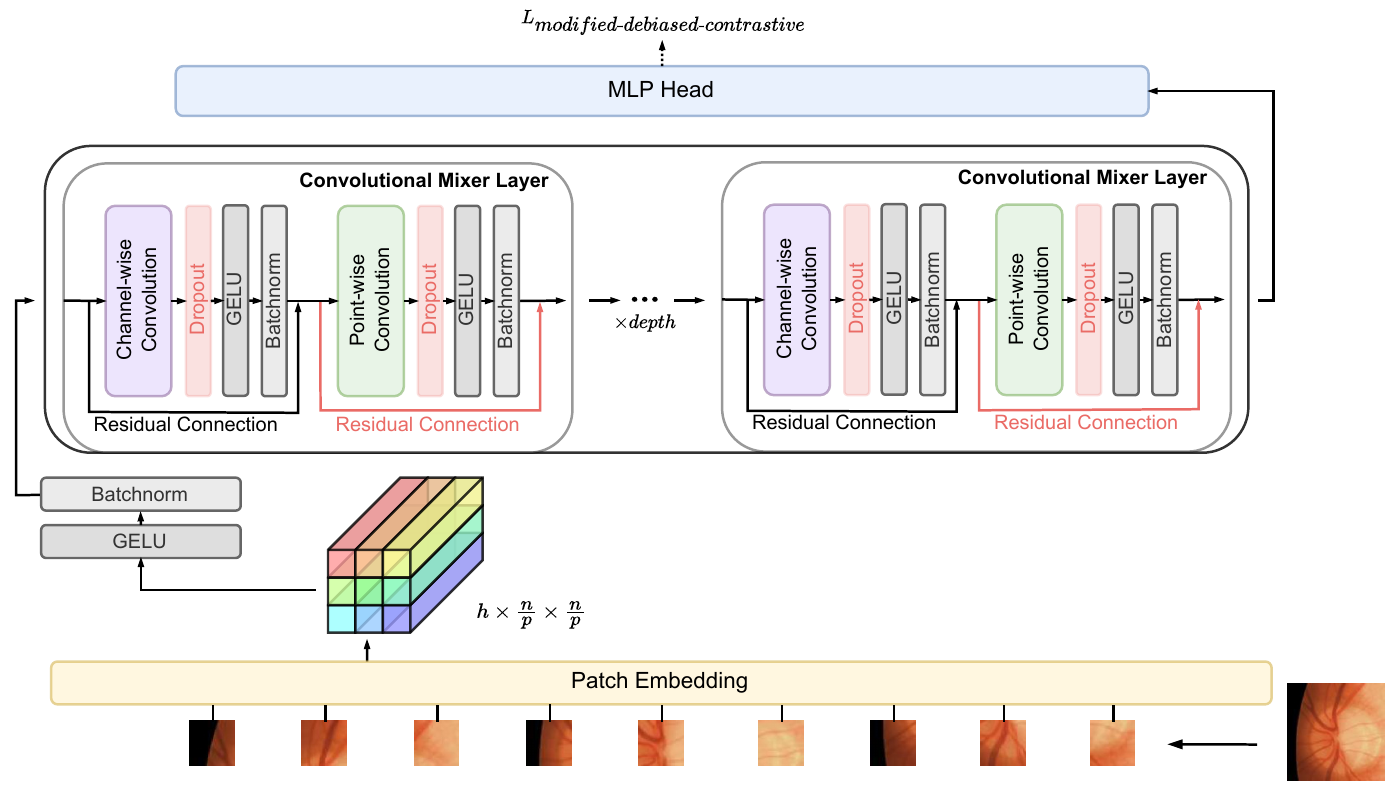}
    \caption{Illustration of the encoder architecture based on ConvMixer~\cite{trockman2022convmixer}. First an input image is divided into relatively small patches which are then processed in mixing layers, consisting of subsequent channel-wise and point-wise convolutions. In contrast to the original ConvMixer the convolutions are followed by a dropout layer in order to prevent over-fitting and to extract more salient features. Following the idea of residual networks~\cite{he2016resnet} we added another residual connection (in red) to enable the point-wise convolution more flexibility during training and to increase the hypothesis space even further. Applying consecutive mixing layers allows the encoder to learn the global structure of the images and can be seen as a form of self attention.}
    \label{fig:ConvMixer}
\end{figure*}


\section{Method}

Fig.~\ref{fig:method} shows our proposed method.
An encoder is used to learn common representations, which are then processed by two parallel networks: the \emph{representation network} and the \emph{deep divergence clustering network}.
The representation network learns representations using our modified debiased contrastive loss, while the clustering network ensures that the learned representations cluster faithfully.
Each of these two modules comes with its own loss which is mixed following a parameter $\gamma$ into the main loss used for training:
\begin{align}
    \mathcal{L}_{MTL} = \mathcal{L}_{deb}^{mod} + \gamma \cdot \mathcal{L}_{clustering}
\end{align}

\subsection{Image encoding}
Before clustering and representation learning, images are encoded to a common representation in an embedding space $Z$ by an encoder $f$.
We used ConvMixer, an isotropic vision model that operates on patches, in order to preserve some local structure within each part of an image~\cite{trockman_patches_2021}.
The input image is first divided into patches of size $p_s$ and dimension $d_h$, which are then fed into series of convolutional mixing blocks consisting of subsequent depth-wise convolutions, in order to mix the spatial structure of the image, and point-wise convolutions, in order to mix channel locations.
To this base architecture we added another residual connection from the output of the depth-wise convolution to the output of the point-wise convolution.
In this sense, our backbone model differs from ~\cite{trockman_patches_2021}, as only one residual connection for spatial awareness was used and thus the original ConvMixer possesses only flexibility in regards to the depth of an image.
By repeated mixing, stacking mixing blocks, an arbitrary large receptive field can be created as distant spatial structures are mixed together the more mixing blocks are used~\cite{trockman_patches_2021}.  

\subsection{Self-supervised Representation Learning}
As shown in Fig.~\ref{fig:method}, our method takes the original image and creates two augmented views using two random transform functions $t_1,t_2$.
The augmented views are generated by applying random cropping, resizing and random Gaussian blurs sequentially on an image twice~\cite{chen_simple_2020}, where the resizing is meant to bring the dimensionality of the cropped image back to its input dimensions.
The encoder network $f$ then projects a sample image $x_i^{(j)}$ onto the common embedding space before a further MLP head $h$ gives the final representations used with our modified contrastive loss.
Previous work~\cite{} has shown that performance in downstream tasks benefits from using intermediate representations rather than those directly used for contrastive learning.

Specifically, from an input image $x_i$ we derive two representations $z_i^{(k)}:=h(f(t_k(x_i)))$, $k\in\{1,2\}$, thus generating $2N$ representations from a mini-batch $B$ with cardinality $N$.
For convenience, we group the representations of all samples $j\neq i$ into $\bar{Z}_i=\{ z_j^{(k)} | j\neq i, k\in\{1,2\} \}$.
The representation of a sample $z_i^{(k)}$ is contrasted via the cosine-similarity $sim(z_i, z_j):=(z_i^\top z_j)(z_i^\top z_i)^{-1}(z_j^\top z_j)^{-1}$ with temperature $\tau$ to the representations of all other samples, defining an average distance:
\vspace{-6pt}
\begin{align}
S_i^{(k)}&=
\frac{1}{|\bar{Z}_i|}\sum_{z\in\bar{Z}_i}
\exp(sim(z_i^{(k)}, z)/\tau)
\end{align}
Due to the absence of training labels, $\bar{Z}_i$ can contain representations of samples belonging to the same category as sample $i$, leading to sampling biases in the regular contrastive learning setting.
The contrastive loss proposed by~\cite{chuang_debiased_2020} solves this problem, but is still vulnerable to the issue of under-clustering of minority classes we discussed in the introduction.

To tackle this problem, we introduce a smoothing term $\lambda$ that softens the impact of the distance $D_i$ of the representations of sample $i$ with those in $\bar{Z}_i$:
\begin{align} \label{loss_con}
\mathcal{L}_{deb,i}^{mod} = -2 \log \frac{\exp(sim(z^{(1)}_i, z^{(2)}_i)/\tau)}{\exp(sim(z^{(1)}_i, z^{(2)}_i)/\tau) + (1+D_i)^\lambda}
\end{align}
where 
\begin{align*}
D_i
&=\sum_{k=1}^2
\max
\bigg\{
\exp(-1/\tau), \\
&\quad\quad\quad
\frac{1}{1-\tau^+}
\bigg(
S_i^{(k)} - \tau^+
\exp(sim(z_i^{(1)},z_i^{(2)})
\bigg)
\bigg\}
\end{align*}
and $\tau^+$ is the prior probability that a sample belongs to the same class of $x$.
Through $\lambda$  we control the emphasis on under-clustering, since an excessive amount of negative samples forces the formation of different clusters within the same category~\cite{wang_solving_2021}, thereby tackling the problem of \emph{within cluster imbalance}.

Applying Eq.~\ref{loss_con} on a mini-batch $B$ results in the loss for the representation learning network:
\begin{align} \label{loss_mod}
  \mathcal{L}_{deb}^{mod}= \frac{1}{2N}\sum_{i=1}^{N}\mathcal{L}_{deb,i}^{mod}
\end{align}

\subsection{Self-Supervised Debiased Clustering} 
The purpose of the clustering network is to refine and improve the learned contrastive representations such that they cluster properly.
We assume that the dataset consists of a long tailed distribution of images over a known number of $K$ categories.
The clustering is refined by pushing the soft label assignments $q_{ij}$ towards the target distribution $P$ by matching them via KL-divergence~\cite{xie2016unsupervised}:
\begin{align} \label{kl}
    \mathcal{L}_{clustering}=KL(P||Q)= \sum_{i=1}^{N}\sum_{j_1}^{K} p_{ij} log \frac{p_{ij}}{q_{ij}}
\end{align}

The embeddings $z_i:=f_(x_i)$ generated by the encoder are subject to a $K$-Means clustering algorithm, where the similarity of an embedded image to the cluster center $\mu_j$, with $j\in K$ is measured by the Student's t-distribution~\cite{maaten_learning_2009}:
\begin{align} \label{q}
    q_{ij}=\frac{(1+||z_i - \mu_j||_2^2/\alpha)^{-\frac{\alpha +1}{2}}}{\sum_{l=1}^{K}(1+||z_i - \mu_l||_2^2/\alpha)^{-\frac{\alpha +1}{2}}}
\end{align}
Where $\alpha$ denotes the degree of freedom and will be set to 1 throughout~\cite{maaten_learning_2009}.
The cluster centers are initialized by performing standard K-Means clustering on the embeddings \cite{xie2016unsupervised}.
Raising the soft label assignments $q_{ij}$ to the second power and normalizing it by the cluster frequencies $u_j=\sum_{i=1}^{N} q_{i,j}$ generates an auxiliary target distribution $P$ for self-supervision \cite{xie2016unsupervised}.
The single elements of $P$ can be computed by:
\begin{align} \label{p}
    p_{ij}=\frac{q_{ij}^2/u_j}{\sum_{j=1}^{K}q_{ij}^2/u_j}
\end{align}
$p_{ij}$ sharpens $q_{ij}$ and due to the normalization reduces bias from imbalanced clusters, while forced to learn the soft label assignments with high confidence~\cite{zhang_supporting_2021}.

\section{Implementation} ~\label{sec:impl}
We train our framework for multiple pretext tasks learning. We follow standard protocols by self-supervised learning for empirical analysis and evaluate the learned representation of our model by classification, semi-supervised, as well as image clustering tasks on different datasets and different computer vision tasks.

\noindent\textbf{Datasets and tasks}
We consider a group of standard datasets in two scenarios. First, relatively balanced data, and secondly, imbalanced setting. For the balanced settings, we used CIFAR-10 \cite{cifar}, CIFAR-100 \cite{cifar}, and Glaucoma-1, which is the labeled subset of the collection of human retinal images used by \cite{diaz2019retinal}. This dataset is composed of 2,397 samples where 956 are diagnosed with glaucoma, and 1421 with no-glaucoma.

For the imbalanced analysis, we performed our experiments on imbalanced CIFAR-10/100 and Glaucoma-2~\cite{orlando2020refuge}. We followed the suggested setting by Cao et al.~\cite{cao2019learning} for the imbalanced CIFAR10/100 in the long-tailed setting and 1:100 as an imbalanced ratio. The Glaucoma-2 contains retina microscopic images in two classes of healthy and Glaucoma with an imbalanced ratio of 1:9. The dataset was released at the REFUGE-2 challenge \cite{orlando2020refuge}, part of MICCAI conference 2020/21. The ISIC-2018 dataset was released at MICCAI 2018 as a challenge dataset and it contains 7 different skin lesions where the imbalance ratio is 1:1:3:5:10:11:69.


\noindent\textbf{Image augmentation}
Similar to SimCLR~\cite{chen_simple_2020}, the random transformation function $\bm T$ applies a combination of image cropping, color jittering, grayscale, horizontal flip, and Gaussian blur. We apply crops with a random size ranging from $0.08$ to $1.0$ of the original area and a random aspect ratio from $3/4$ to $4/3$ of the original aspect ratio. The cropped part is then resized to the original size. We also apply horizontal flipping with a probability of $0.5$ and apply grayscale with a probability of $0.2$ as well as color jittering with a probability of $0.8$ and a configuration of $(0.4, 0.4, 0.4, 0.1)$. Gaussian blur is applied with a kernel size of $10\%$ of the original image size, a $\sigma$ randomly sampled from $[0.1, 2.0]$, and a probability of $0.5$.

\noindent\textbf{Deep debiased representation architecture}
First the encoder takes an input image and divides it into quadratic patches of size $p$. These patches are then fed into a sequence of $d$ Convolutional Mixer Layers which consist of two types of separable convolutions: depth-wise convolutions and point-wise convolution. Here, we modified the supervised ConvMixer architecture~\cite{trockman2022convmixer} 1) by making a network self-supervised, 2) adding a new residual connection in order to create further flexibility for the depth awareness of the network, and 3) adding the dropout layers. As depicted in Fig. \ref{fig:ConvMixer}, the Convolutional Mixing Layers contain channel-wise followed by point-wise convolutions. After each convolution a dropout layer is implemented with a dropout rate followed by a GELU activation layer (Gaussian error linear unit) and a batch normalization. Moreover, each convolution is endowed with corresponding residual connections. Finally, the embeddings are generated by a linear projection of the last Convolutional Mixing Layer.

The experiment is conducted on $p=2$ pixel patches and $d=8$ mixing blocks. Both convolutions have an equal number of channels with 256. The depth-wise convolution uses a kernel size of 7. The linear projection contains two fully connected layers. The first one projects onto 512 hidden neurons and the second projects onto a 128 dimensional embedding space. 

We also train our deep unsupervised clustering network with the same architecture as described above (see Fig. \ref{fig:method}).  

\noindent\textbf{Optimization}
Following \cite{chuang_debiased_2020}, we used the Adam optimizer \cite{kingma2014adam} with a learning rate of $10^-3$, $\beta_1 = 0.9$, $\beta_2 = 0.999$, and a weight decay of $10^-6$. Accordingly, the temperature hyperparameter $\tau$ was set to $0.5$ and the positive class prior $\tau^+$ to $0.1$. 
The emphasis on negative samples $\gamma_{mod}$ and the weight of the clustering loss $\gamma$ were set to 2 and 5, respectively. They were tuned by grid search on $\{(\gamma, \gamma_{mod}): \gamma$ $\epsilon$ $\{0.1, 1, 5\}$, $\gamma_{mod}$ $\epsilon$ $\{2, 3, 5\}\}$. 
We chose a dropout rate of $0.04$ and different mini-batch sizes depending on the image size. For CIFAR-10/-100, we used a size of $256$ and for the medical datasets a size of $32$. 
All the models were trained for 500 epochs on a single GPU (Tesla A100) with 40 GB of memory.

\section{Experiments and Results} 

\begin{figure}[!t]
    \centering
    \includegraphics[width=0.49\textwidth]{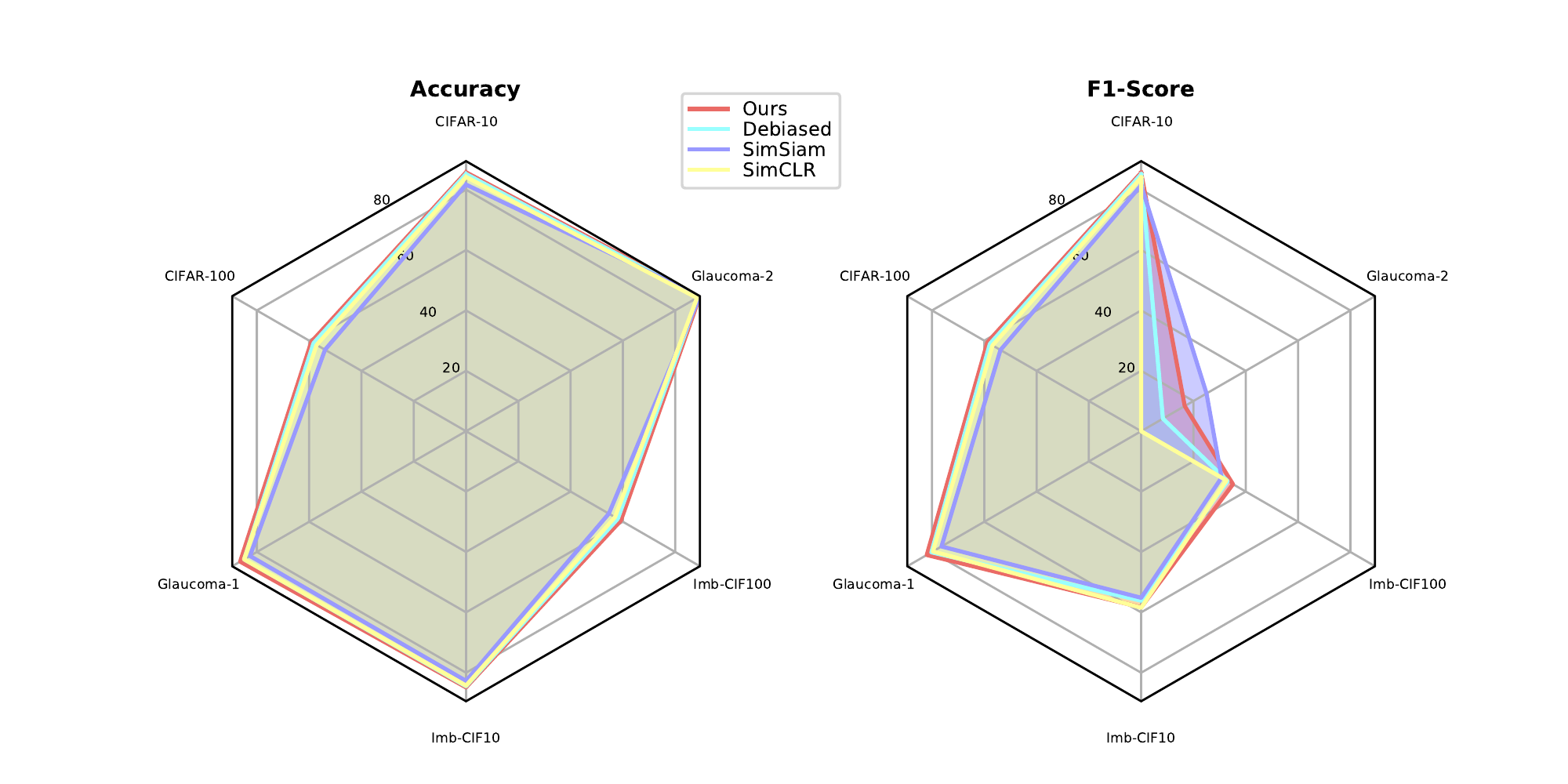}
    \caption{Comparison of the proposed method with baselines under \underline{linear evaluation} measured by top-1 accuracy (Left) followed by F1-score (Right).}
    \label{fig:lineareval}
\end{figure}

\noindent\textbf{Linear Evaluation} 
Following the standard linear evaluation protocol~\cite{chen_simple_2020, chuang_debiased_2020}, the representations of the backbone network were fixed and the projection head was removed after training. Then, a linear classifier on top was trained using the cross entropy loss for 100 epochs with a batch size of 512. Table \ref{table:lineareval:balanced} shows the achieved performance on relative balanced dataset. Considering the imbalance in datasets, we report the Binary or Macro F1-score depending on the number of classes, in addition to the top-1 accuracy on the respective testing sets in Table \ref{table:lineareval:imb} and Fig.~\ref{fig:lineareval}. 

\begin{table*}
\small
\centering
\caption{Comparison of the proposed method with baselines under \underline{linear evaluation} on relatively balanced dataset and measured by top-1 accuracy (\%) and F1-score, respectively.} 
\label{table:lineareval:balanced}
\begin{tabular}{l| c c c }
\toprule
Method   & Glaucoma-1  & CIFAR-10  & CIFAR-100  \\
\midrule
SimCLR~\cite{chen_simple_2020}   & 84.86 $\pm$ 1.37/79.63 $\pm$ 1.80 &  84.20 $\pm$ 0.06 /84.03 $\pm$ 1.4 & 57.15 $\pm$ 0.11/56.84 $\pm$ 0.05 \\
SimSiam~\cite{chen2021exploring} & 82.64 $\pm$ 0.98 /76.36 $\pm$ 1.77 & 81.68 $\pm$ 0.57/ 81.45 $\pm$ 0.59 & 53.93 $\pm$ 0.40 /53.8 $\pm$ 0.43 \\
Debiased~\cite{chuang_debiased_2020} & 84.86 $\pm$ 0.20/80.22 $\pm$ 0.30  & 85.19 $\pm$ 0.41 /85.09 $\pm$ 0.39  & 58.32 $\pm$ 0.55 /57.85 $\pm$ 0.63  \\ \midrule
Our method & \textbf{86.39  $\pm$ 0.40 /81.99 $\pm$ 0.15}  & \textbf{85.74 $\pm$ 0.34 /85.64 $\pm$ 0.29}  & \textbf{59.18 $\pm$ 0.37/58.83 $\pm$ 0.49} \\ 
\bottomrule
\end{tabular}
\end{table*}

\begin{table*} 
\small
\centering
\caption{Comparison of the proposed method with baselines on imbalanced dataset under linear evaluation measured by top-1 accuracy (\%) and F1-score, respectively. } 
\label{table:lineareval:imb}
\begin{tabular}{l| c c c }
\toprule
Method     & Glaucoma-2  & Imb-CIFAR 10 & Imb-CIF100 \\
\midrule
SimCLR~\cite{chen_simple_2020} & 88.39 $\pm$ 0.96 /0.00 $\pm$ 0.5 & 84.26 $\pm$ 0.34/58.46 $\pm$ 0.44 & 56.95 $\pm$ 1.10/32,60 $\pm$ 1.03 \\
SimSiam~\cite{chen2021exploring} & 89.04 $\pm$ 0.91 /23.02 $\pm$ 1.43 & 82.69 $\pm$ 0.17 /55.17 $\pm$ 0.08 & 54.62 $\pm$ 0.59/30.75 $\pm$ 0.33 \\
Debiased~\cite{chuang_debiased_2020} & 88.89 $\pm$ 0.92/27.07 $\pm$ 4.3& 84,47 $\pm$ 0.91 /56.14 $\pm$ 0.14 & 58,03 $\pm$ 0.43/33.08 $\pm$ 0.53 \\ \midrule
Our method  & \textbf{89.44 $\pm$ 0.18/60.17 $\pm$ 0.3} & \textbf{84.85 $\pm$ 0.83/58.33 $\pm$ 2.77} & \textbf{59.22 $\pm$ 1.33/35.12 $\pm$ 1.20} \\ 
\bottomrule
\end{tabular}
\end{table*}

\begin{figure}[!t]
    \centering
    \includegraphics[width=0.49\textwidth]{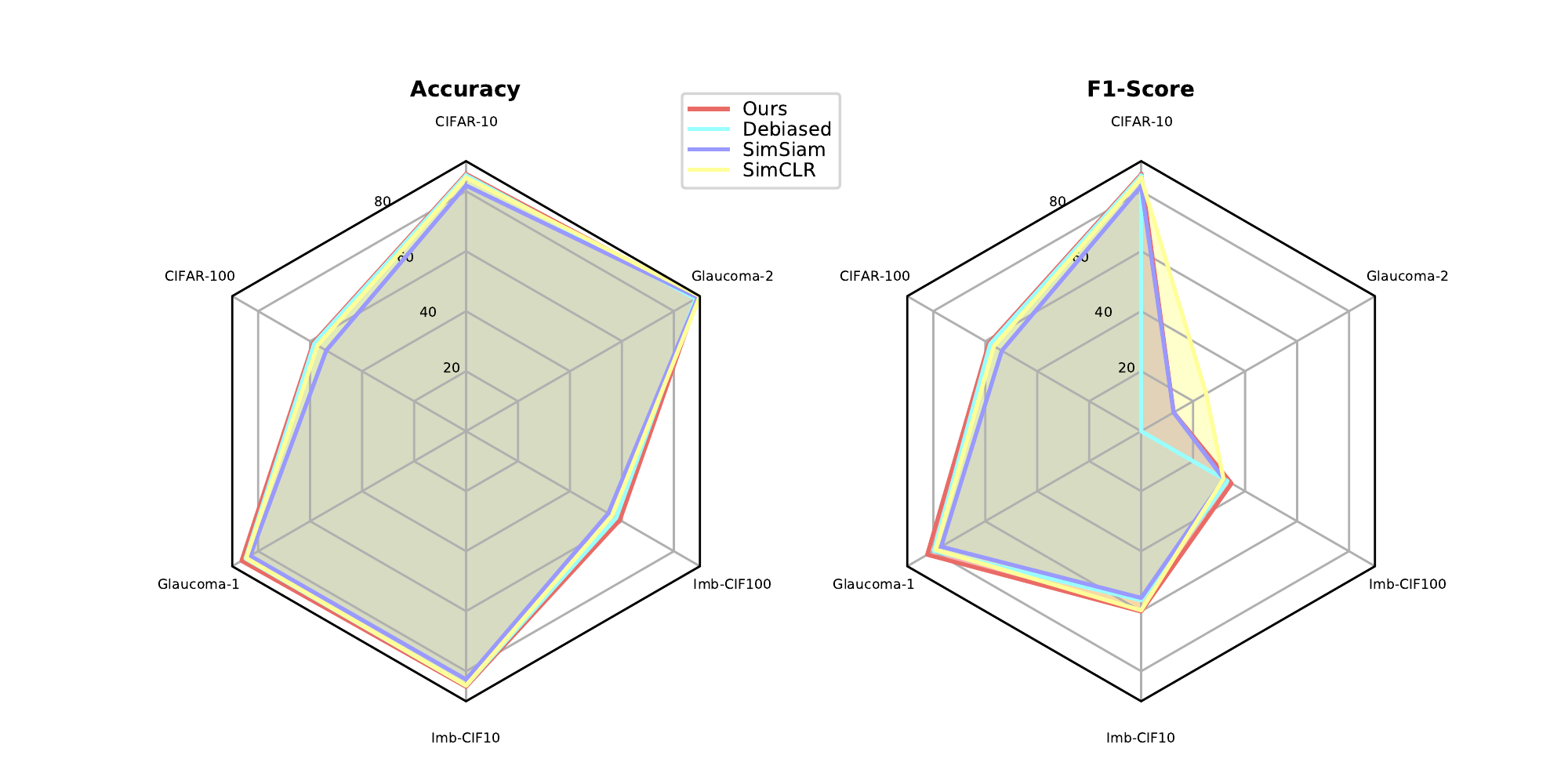}
    \caption{Comparison of the proposed method with baselines under \underline{semi-supervised evaluation} measured by top-1 accuracy (Left) followed by F1-score (Right).}
    \label{fig:semisup}
\end{figure}

\noindent\textbf{Semi-Supervision} \space 
For testing the models in a semi-supervised scenario, we train the linear classifier using only 10\% of the training data. Table \ref{table:semieval:balanced}, \ref{table:semieval:imb}, and Figure \ref{fig:semisup} shows the top-1 accuracy and the F1-score for balanced and imbalanced dataset respectively. 

\begin{table*}
\small
\centering
\caption{Comparison of the proposed method with baselines on balanced dataset under \underline{semi-supervised evaluation} measured by top-1 accuracy (\%) and F1-score, respectively.} 
\label{table:semieval:balanced}
\begin{tabular}{l| c c c }
\toprule
Method & Glaucoma-1 & CIFAR-10 & CIFAR-100 \\
\midrule
SimCLR~\cite{chen_simple_2020} & 84.59 $\pm$ 0.98/79.10 $\pm$ 1.27 & 84.50 $\pm$ 0.69/84.37 $\pm$ 0.69 & 57.18 $\pm$ 0.28/56.67 $\pm$ 0.30 \\
SimSiam~\cite{chen2021exploring} & 83.06 $\pm$ 0.00/76.97 $\pm$ 0.74 & 81.88 $\pm$ 0.42/81.71 $\pm$ 0.42 & 53.80 $\pm$ 0.52/53.63 $\pm$ 0.54 \\
Debiased~\cite{chuang_debiased_2020} & 84.86 $\pm$ 1.37/79.77 $\pm$ 1.99 & 85.27 $\pm$ 0.37/85.14 $\pm$ 0.39 & 58.41 $\pm$ 0.18/58,11 $\pm$ 0.23 \\  \midrule
Our method  & \textbf{86.30 $\pm$ 1.17/82.32 $\pm$ 1.26} & \textbf{85.72 $\pm$ 0.41/85.62 $\pm$ 0.43} & \textbf{58.82 $\pm$ 0.01/58.63 $\pm$ 0.13} \\ 
\bottomrule
\end{tabular}
\end{table*}

\begin{table*}
\small
\centering
\caption{Comparison of the proposed method with baselines on imbalanced dataset under semi-supervised evaluation measured by top-1 accuracy (\%) and F1-score, respectively.} 
\label{table:semieval:imb}
\begin{tabular}{@{}l@{\hspace{.1cm}}|@{\hspace{.1cm}}c@{\hspace{.4cm}}c@{\hspace{.4cm}}c@{\hspace{.4cm}}c@{}}
\toprule
Method  & ISIC & Glaucoma-2 & Imb-CIF10 & Imb-CIF100 \\
\midrule
SimCLR~\cite{chen_simple_2020} & 68.55 $\pm$ 0.04/38.62 $\pm$ 0.00 & 88.00 $\pm$ 0.00/55.00 $\pm$ 0.00 & 84.77 $\pm$ 0.14/59.71 $\pm$ 0.64 & 56.25 $\pm$ 0.63/31.70 $\pm$ 0.34\\
SimSiam~\cite{chen2021exploring} & 61.31 $\pm$ 0.08/31.21 $\pm$ 0.01 & 89.17 $\pm$ 1.18/42.50 $\pm$ 17.68 & 82.75 $\pm$ 0.14/55.56 $\pm$ 3.73 & 54.74 $\pm$ 1.23/31.83 $\pm$ 1.47\\
Debiased~\cite{chuang_debiased_2020}& 62.43 $\pm$ 0.05/33.11 $\pm$ 0.00 & 88.33 $\pm$ 0.00/52.06 $\pm$ 0.00& 84.79 $\pm$ 0.11/56.34 $\pm$ 3.73 & 57.79 $\pm$ 0.76/33.20 $\pm$ 0.21\\  \midrule
Our method  & \textbf{73.29 $\pm$ 0.04/67.12 $\pm$ 0.01} & \textbf{89.19 $\pm$ 1.18/72.50 $\pm$ 0.68} & \textbf{85.17 $\pm$ 0.48/60.09 $\pm$ 1.41} & \textbf{59.20 $\pm$ 0.37/34.79 $\pm$ 0.43}\\ 
\bottomrule
\end{tabular}
\end{table*}

\noindent\textbf{Unsupervised Clustering} \space 
Clustering is examined by running the K-Nearest Neighbor algorithm (KNN) with the features of the pretrained encoder. We use a relatively small neighborhood size of 20, considering possible small classes in imbalanced datasets, and report the outcomes in  \hyperref[table:clusteringeval]{Table \ref{table:clusteringeval}}. 

\begin{table*} [!t]
\centering
\small
\caption{Comparison of the proposed method with the baseline on image clustering measured by top-1 accuracy (\%) and F1-score.} 
\label{table:clusteringeval}
\begin{tabular}{l| c c c c c c c}
\toprule
Method  & ISIC & Glaucoma-1 & Glaucoma-2 & CIFAR-10 & Imb-CIFAR-10 & CIFAR-100 & Imb-CIFAR-100 \\
\midrule
SimCLR~\cite{chen_simple_2020}      & 48.50/14.46 & 81.17/80.41 & 90.00/40.00 & 78.11/77.97 & 79.18/49.63 & 47.84/47.78 & 45.38/27.14 \\
SimSiam~\cite{chen2021exploring}     & 47.64/14.35 & 80.83/76.45 & 86.67/33.33 & 75.21/75.17 & 74.82/46.70 & 43.01/42.79 & 41.30/23.23 \\
Debiased~\cite{chuang_debiased_2020}    & 47.70/14.55 & 83.33/79.45 & 93.33/60.00 & 80.70/80.59 & 80.31/50.37 & 49.08/49.01 & 46.03/25.75 \\  \midrule
Our method  & \textbf{49.30/15.42} & \textbf{84.50/78.05} & \textbf{96.32/68.36} & \textbf{80.93/80.86} & \textbf{81.51/63.43} & \textbf{49.89/49.09} & \textbf{46.50/25.52} \\
\bottomrule
\end{tabular}
\end{table*}

\section{Ablation Study}
We further investigate the following aspects of our approach in multiple ablation analysis: (1) analysis of clustering loss and hyperparameters, (2) the impact of modified ConvMixer networks, and (3) robustness of our algorithm for training on imbalanced data distribution.

\noindent\textbf{Network architecture} We consider different network architecture as an encoder architecture such as ResNet-18~\cite{he2016resnet} and CoAtNet~\cite{dai2021coatnet}.
ResNet-18 composed of the 17 convolutional layers ~\cite{he2016resnet} and it addressed the problem of vanishing gradients in deep networks by residual connections. The network learns the identify function of the previous convolutional layer, which makes the training of deep convolutional neural networks (CNN) more efficient and gives the ability to compress the model, reducing the need for neural architecture search. CoAtNet unites depth-wise convolution and relative self attention~\cite{dai2021coatnet}. The main idea is to keep the desirable properties such as the generalizability of CNNs and the high model capacity of transformers. This is achieved by stacking convolutional and transformer blocks. The convolutional block consisting of an expansion, depth-wise convolution and following compression to the input size. The Transformer block starts with a relative attention layer, which can be skipped through a residual connection same as the whole convolutional block and a subsequent multi layer perceptron. The input image is divided into patches to enable global attention–\cite{dosovitskiy2020image} and then fed into a combination of stacked convolutional and transformer blocks before being projected by a fully connected layer. Table~\ref{table:arch} shows and compares the impact of encoder architecture.

\begin{table*} [!t]
\small
\centering
\caption{Top-1 accuracy (\%) and F1 score for image clustering evaluation where all the networks are trained with our proposed loss function.}
\label{table:arch}
\begin{tabular}{l l l l l l l l }
\toprule
Encoder/dataset & CIFAR10 & CIFAR100 & Imb-CIF-10 & Imb-CIF-100 & Glucoma-1 & Glucoma-2 & ISIC \\
\midrule
ResNet & 78.66/78.64 & \textbf{56.28/56.42} & \textbf{83.62/67.33} & \textbf{50.07/27.39} & 84.17/80.81 & 91.23/46.15 & 48.30 \\
CoAtNet & 74.78/74.66 & 43.56/43.32 & 76.92/47.25 & 41.48/23.10 & 83.11/80.13 & 90.0/40.0 & \textbf{50.23} \\
ConvMixer & \textbf{80.93/80.86} & 49.89/49.09 & 81.51/63.43 & 46.50/25.52 & \textbf{84.50/80.05} & \textbf{96.32/68.36} & 49.30 \\
\bottomrule
\end{tabular}
\end{table*}

\noindent\textbf{Clustering} We examine 
Clustering was performed with the K-Nearest Neighbor algorithm (KNN). We used a neighborhood size of 20 and report the top-1 accuracy and the (Binary / Macro) F1-score in \hyperref[table:clusteringeval]{Table \ref{table:clusteringeval}}. 


\noindent\textbf{Robustness against imbalanced data distribution}
Since the success of deep learning methods in a variety of downstream task on balanced data, researchers steadily shifted their focus on solutions for imbalanced data problems, as they are a more accurate portray of real world data~\cite{wang_contrastive_2021}. In a supervised setting, the problem can be tackled by the use of data balancing techniques such as down sampling the majority class~\cite{buda_systematic_2018,drummond_c45_2003} or up sampling the infrequently observed classes~\cite{buda_systematic_2018,sarafianos_deep_2018}, both of which aims to produce a more balanced data set during the training process~\cite{mahajan_exploring_2018}. The effectiveness of data balancing is intuitively given, nevertheless, it has to be said that those methods come with their own set of downsides. down sampling can lead to a loss of information within the majority class and up sampling brings the risk of over-fitting the minority class, as no more additional information is generated~\cite{more_survey_2016}. Other solutions modify the loss function, such that it can be defined as cost sensitive. loss re-weighting techniques puts a higher weight on minority classes and a lower weight on majority classes, effectively~\cite{khan_cost_2017,cui_class-balanced_2019}. The single re-weighting methods mainly differ in how and what weights are assigned to which class~\cite{wang_contrastive_2021}.

\section{Conclusion} \label{sec:conclusion}
In this paper, we introduce a new self-supervised multi-task learning framework which learns simultaneously unsupervised representation and perform image clustering. We explored a self-supervised embedding trained jointly for clustering and debiased representation learning is more robust for dataset with imbalanced distribution. Moreover, image clustering benefits of this of self-supervised embedding. In this study, we addressed problem of under-clustering which is due to imbalanced distribution by re-designing the previous debiased contrastive loss. Then, networks are optimized end-to-end using our modified debiased contrastive loss and KL-divergence loss. We improved over previous methods for deep clustering, deep self-supervised, and semi-supervised learning. Empirical experiments demonstrate the efficacy of the proposed method on standard benchmarks as well as imbalanced clinical datasets. 

\paragraph{Acknowledgments:} We would like to thank Priyank Jaini for valuable feedback on the manuscript. We also would like to acknowledge that E. D. is supported by the Helmholtz Association under the joint research school ``Munich School for Data Science - MUDS'' (Award Number HIDSS-0006), M. R. and B. B. were supported by the Bavarian Ministry of Economic Affairs, Regional Development and Energy through the Center for Analytics – Data – Applications (ADA-Center) within the framework of BAYERN DIGITAL II (20-3410-2-9-8), and M. R. and B. B. were supported by the German Federal Ministry of Education and Research (BMBF) Munich Center for Machine Learning (MCML). 

\newpage

{\small
\bibliographystyle{ieee_fullname}
\bibliography{egbib}
}

\end{document}